\documentclass[10pt,twocolumn,letterpaper]{article}

\usepackage[pagenumbers]{cvpr} 

\usepackage{url}

\usepackage{graphicx}
\usepackage{kotex}
\usepackage{multirow}
\usepackage{multicol}
\usepackage[table,dvipsnames]{xcolor}
\usepackage{caption}
\usepackage{wrapfig}
\usepackage{float}

\usepackage{booktabs}
\usepackage{enumitem}
\usepackage{amsmath}
\usepackage{amsfonts}
\usepackage{nicefrac}

\usepackage{adjustbox}
\usepackage{soul}
\usepackage{pifont}
\usepackage{array}
\usepackage{makecell}

\usepackage{times}
\usepackage{helvet}
\usepackage{courier}
\usepackage{arydshln}

\newcommand{\conf}{\scriptsize\textcolor{Gray}}

\newcommand{\figref}[1]{Fig.~\ref{#1}}
\newcommand{\tabref}[1]{Tab.~\ref{#1}}
\newcommand{\secref}[1]{Sec.~\ref{#1}}

\definecolor{cvprblue}{rgb}{0.21,0.49,0.74}
\usepackage[pagebackref,breaklinks,colorlinks,allcolors=cvprblue]{hyperref}

\def\paperID{10146} 
\def\confName{CVPR}
\def\confYear{2026}

\title{CVA: Context-aware Video-text Alignment for Video Temporal Grounding}

\author{
    Sungho Moon\textsuperscript{*}\quad 
    Seunghun Lee\textsuperscript{*}\quad 
    Jiwan Seo\quad 
    Sunghoon Im\textsuperscript{†} \\
    Daegu Gyeongbuk Institute of Science and Technology (DGIST), Republic of Korea \\
    {\tt\small \{byeol3325, lsh5688, eccaron, sunghoonim\}@dgist.ac.kr} \\ [-0.3em]
    {\small \url{https://byeol3325.github.io/projects/CVA/}} 
}

\begin{document}
\maketitle
\begin{abstract}
We propose \textbf{Context-aware Video-text Alignment (CVA)}, a novel framework to address a significant challenge in video temporal grounding: achieving temporally sensitive video-text alignment that remains robust to irrelevant background context. 
Our framework is built on three key components. First, we propose \textbf{Query-aware Context Diversification (QCD)}, a new data augmentation strategy that ensures only semantically unrelated content is mixed in. It builds a video-text similarity-based pool of replacement clips to simulate diverse contexts while preventing the ``false negative" caused by query-agnostic mixing. Second, we introduce the \textbf{Context-invariant Boundary Discrimination (CBD) loss}, a contrastive loss that enforces semantic consistency at challenging temporal boundaries, making their representations robust to contextual shifts and hard negatives. Third, we introduce the \textbf{Context-enhanced Transformer Encoder (CTE)}, a hierarchical architecture that combines windowed self-attention and bidirectional cross-attention with learnable queries to capture multi-scale temporal context.
Through the synergy of these data-centric and architectural enhancements, CVA achieves state-of-the-art performance on major VTG benchmarks, including QVHighlights and Charades-STA. Notably, our method achieves a significant improvement of approximately 5 points in Recall@1 (R1) scores over state-of-the-art methods, highlighting its effectiveness in mitigating false negatives.

\end{abstract}

\begin{figure}[!t]
    \centering
    \includegraphics[width=1\linewidth]{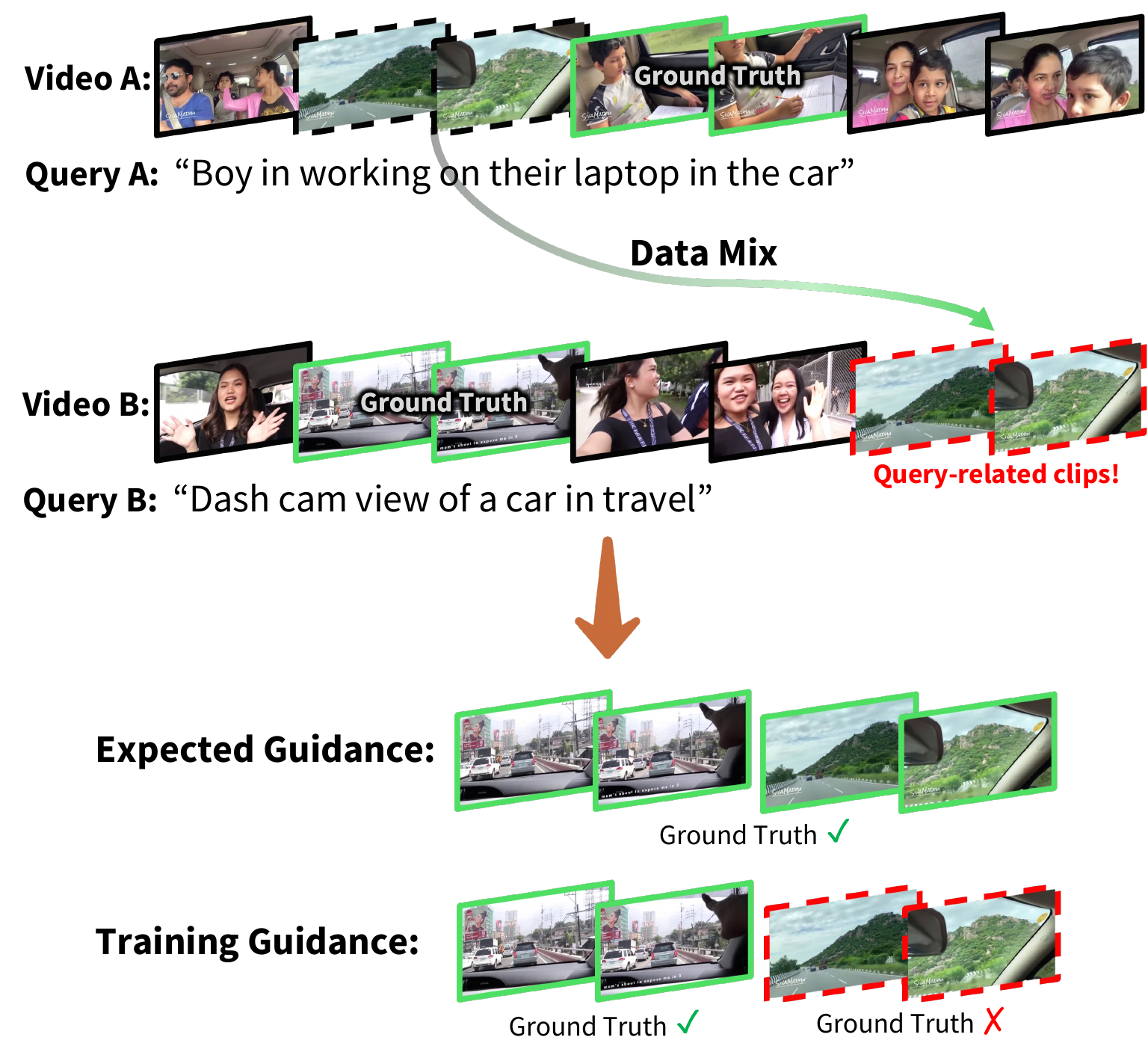}
    \caption{Problem of previous context diversification \cite{zhou2025devil}.}
    \label{fig:motiv}
\end{figure}

\section{Introduction}

The rapid growth of online video platforms, such as YouTube and TikTok, has led to an unprecedented surge in the volume and diversity of video content. This abundance poses a major challenge in enabling users to efficiently browse and retrieve content that aligns with their specific interests. Consequently, the field of Video Temporal Grounding (VTG) has garnered substantial research attention, aiming to bridge the gap between human language and video content. VTG encompasses key tasks such as Video Moment Retrieval (VMR) \cite{gao2017tall, anne2017localizing}, which localizes segments corresponding to text queries, and Highlight Detection (HD) \cite{sun2014ranking, yao2016highlight}, which identifies the most salient segments. Following the pivotal introduction of a DETR-based architecture \cite{carion2020end} and the QVHighlights dataset by Lei \etal. \cite{lei2021Qvhighlights}, recent research has increasingly focused on enhancing fine-grained video-text alignment.

To advance this goal, prior work has explored synergistic learning of multiple temporal tasks \cite{liu2022umt, lin2023univtg, sun2024tr, xiao2024bridging}, input-dependent query generation \cite{liu2022umt, jang2023knowing}, and negative pair training strategies \cite{moon2023query, jung2025background}. Despite these advancements, a fundamental issue was recently identified by \cite{zhou2025devil}: models tend to learn spurious correlations, overly associating text queries with static backgrounds rather than the target temporal dynamics. To mitigate this, they introduce a content mixing augmentation that replaces background clips with content from other videos. This breaks the link between actions and their original backgrounds and encourages models to focus on the true moment-level semantics.

While effective, content mixing remains query-agnostic, as the replacement clips are sampled without regard for their semantic relevance to the text query. This can generate \textit{false negatives} when semantically related clips are mistakenly treated as negative examples, as shown in \figref{fig:motiv}. To address this limitation, we propose \textbf{Query-aware Context Diversification (QCD)}, an advanced data augmentation strategy that ensures only semantically unrelated clips are used for replacement by analyzing video-text relevance with pre-trained CLIP features. QCD also preserves the immediate temporal context surrounding the ground-truth (GT) moment, recognizing its importance for precise localization.

Complementing this augmentation, we further introduce the \textbf{Context-invariant Boundary Discrimination (CBD) loss}, which explicitly enforces semantic consistency despite the diverse contextual shifts introduced by QCD. By focusing this consistency objective on the temporal boundaries—the regions most critical and challenging for precise alignment—our loss guides the model to learn a highly discriminative and context-invariant representation.

To further strengthen the model's ability to capture temporal structure, we enhance the underlying architecture with the \textbf{Context-enhanced Transformer Encoder (CTE)}, a hierarchical encoder designed to capture multi-scale temporal context. Unlike standard Transformers, CTE employs windowed self-attention to model local patterns, combined with bidirectional cross-attention with learnable queries to refine and aggregate contextual information. 

Through the synergistic combination of these components, our model achieves state-of-the-art performance on major Video Moment Retrieval and Highlight Detection benchmarks. Our contributions are summarized as follows:
\begin{itemize}
    \item We propose \textbf{Query-aware Context Diversification (QCD)}, a novel augmentation strategy that simulates diverse temporal contexts while preventing the false negative issue inherent in query-agnostic mixing.
    \item We introduce the \textbf{Context-invariant Boundary Discrimination (CBD) loss}, a boundary-focused contrastive objective that learns representations invariant to contextual shifts, significantly enhancing localization precision.
    \item We design the \textbf{Context-enhanced Transformer Encoder (CTE)}, a hierarchical architecture that effectively models multi-scale temporal context.
\end{itemize}





%


\section{Related Work}

\subsection{Moment Retrieval}
The task of Video Moment Retrieval (VMR) aims to localize a query-specific moment in untrimmed videos. Early approaches can be broadly categorized into two types: propose-then-rank framework and proposal-free framework. The propose-then-rank framework \cite{gao2017tall, anne2017localizing, ge2019mac, sun2022you, xiao2021boundary} first generates potential proposals and then ranks them based on relevance scores. Proposal-free approaches directly predict the target moments \cite{mun2020local, yuan2019find}, or estimate the probabilities of each frame being a boundary position \cite{ghosh2019excl, zhang2020span}. Recently, the field has converged on end-to-end, query-based models inspired by the Detection Transformer \cite{carion2020end}, pioneered by Moment-DETR \cite{lei2021Qvhighlights}. These models use learnable queries to probe features and directly predict temporal spans. To further enhance video-text alignment, various strategies have been proposed, such as video-specific query generation \cite{liu2022umt, jang2023knowing}, global and local alignment pipelines \cite{sun2024tr, liu2024towards}, and learning with negative pairs \cite{moon2023query, jung2025background, zhou2025devil}. Recently, TD-DETR \cite{zhou2025devil} identified the model’s tendency to over-associate text queries with background frames and proposed a data synthesis strategy to construct dynamic text-grounded contexts.

\subsection{Highlight Detection}

Highlight detection (HD) aims to identify the most salient segments within a video. 
These approaches can be broadly categorized into query-agnostic and query-based methods. 
Early query-agnostic works focused on assessing the intrinsic importance of video clips based solely on visual information to determine salience scores \cite{sun2014ranking, xu2021cross, wei2022learning, badamdorj2022contrastive}.
To better align with specific user interests, query-based HD emerged, incorporating textual information to guide the detection process \cite{dagtas2004multimodal, kudi2017words}. 
This naturally led to the development of advanced multimodal systems that leverage complementary signals. 
Query-based HD shares a close relationship with video moment retrieval (MR). 
While traditionally treated as distinct tasks, recent research has increasingly focused on their joint learning. 
A seminal work in this direction, MomentDETR \cite{lei2021Qvhighlights}, introduced the QVHighlights dataset and a DETR-based framework to facilitate this joint training. 
Building on this, subsequent methods explored various improvements. 
For instance, UMT \cite{liu2022umt} utilized multi-modal content (audio, visual, text) to refine query generation, 
while QD-DETR \cite{moon2023query} enhanced textual understanding via negative relationship learning. 
More recently, models like UVCOM \cite{xiao2024bridging} and TR-DETR \cite{sun2024tr} seek a more comprehensive understanding by integrating the distinct characteristics of both MR and HD.

\begin{figure}[!t]
    \centering
    \includegraphics[width=1\linewidth]{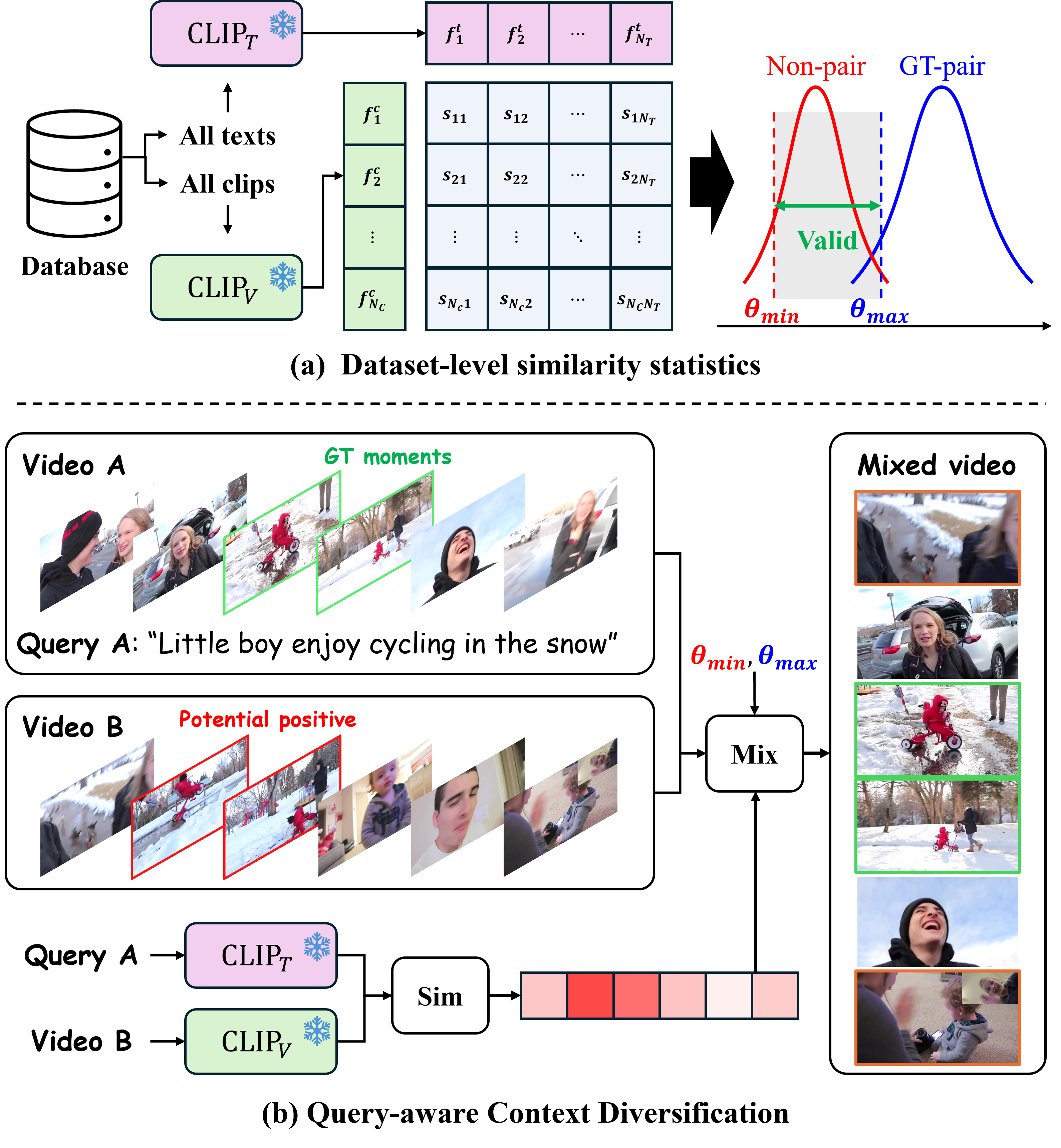}
    \caption{Illustration of our Query-aware Context Diversification.}
    \label{fig:qcd}
\end{figure}

\begin{figure*}[!t]
    \centering
    \includegraphics[width=1\linewidth]{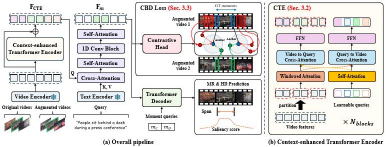}
    \caption{\textbf{Overview of our CVA.} (a) The overall pipeline, which includes the Context-enhanced Transformer Encoder and the Context-invariant Boundary Discrimination module. (b) The detailed architecture of the Context-enhanced Transformer Encoder.}
    \label{fig:cva}
\end{figure*}

\section{Method}

The overall pipeline of CVA is illustrated in \figref{fig:cva}. 
Given an untrimmed video $\mathbf{V}$ and its corresponding text query $\mathbf{Q}$, our goal is to localize temporal moments  $\hat{m} = (\hat{c}, \hat{\sigma})$ corresponding to the query, while also predicting per-clip saliency scores $\mathbf{s} = \{ s_i \}_{i=1}^{L}$. Building upon the transformer-based architecture \cite{lei2021Qvhighlights}, we design a context-aware pipeline capable of simulating and comprehending diverse temporal dynamics. Our approach consists of three key components: (1) a \textit{Query-aware Context Diversification (QCD)} strategy for generating semantically consistent yet contextually diverse training samples; (2) a \textit{Context-enhanced Transformer Encoder (CTE)} designed to capture hierarchical temporal contexts; and (3) a \textit{Context-invariant Boundary Discrimination (CBD) loss} to learn robust boundary representations invariant to contextual shifts.

\subsection{Query-aware Context Diversification}

Given a video $\mathbf{V}_A$ and its corresponding text query $\mathbf{Q}$, we can simulate diverse contexts by replacing the background regions (i.e., non-moment clips) with clips from another video $\mathbf{V}_B$. Let $\mathbf{M} \in \mathbb{B}^{L}$ be a binary preserving mask, where $1$ indicates the clips to preserve (i.e., the GT moment and its adjacent context) and $0$ indicates the clips to be replaced. The mixed video $\mathbf{V}_{\text{mix}}$ is generated as:
\begin{equation}
    \mathbf{V}_{\text{mix}} = \mathbf{M} \odot \mathbf{V}_A + (1 - \mathbf{M}) \odot \mathbf{V}_B,
\end{equation}
where $\odot$ denotes element-wise multiplication. However, a randomly sampled video $\mathbf{V}_B$ may contain clips that are semantically relevant to the query $\mathbf{Q}$. Since the mixed regions are trained as negative (background) segments, this can introduce false negatives and hinder model training. To address this, we propose \textbf{\textit{Query-aware Context Diversification (QCD)}}, a strategy that simulates diverse contexts by exchanging content based on video-text relevance, as illustrated in \figref{fig:qcd}.


To build a query-aware sampling strategy, we pre-compute similarity statistics at the dataset level using a pre-trained CLIP model \cite{radford2021learning}. Let $\mathcal{C} = \{\mathbf{c}_i\}_{i=1}^{N_C}$ be all video clips and $\mathcal{T} = \{\mathbf{t}_j\}_{j=1}^{N_T}$ be the set of unique text queries. For each clip $\mathbf{c}_i$ and text query $\mathbf{t}_j$, we extract their CLIP features using the visual and text encoders as follows: 
\begin{equation}
\begin{gathered}
    \mathbf{F}_{\text{v}} = \{ \mathbf{f}_i^{\text{v}} \}_{i=1}^{L} \in \mathbb{R}^{L \times D_{\text{v}}}, ~\mathbf{f}^{\text{v}}_i = \text{CLIP}_V(\mathbf{c}_i),\\
    \mathbf{F}_{\text{t}} = \{ \mathbf{f}_j^t \}_{j=1}^{N} \in \mathbb{R}^{N \times D_{\text{t}}}, ~\mathbf{f}^{\text{t}}_j = \text{CLIP}_T(\mathbf{t}_j).
\end{gathered}
\end{equation}


We then compute a comprehensive similarity of each element $s_{ij}$ is defined as the cosine similarity as follows:
\begin{equation}
    s_{ij} = \frac{(\mathbf{f}^{\text{v}}_i)^T \mathbf{f}^{\text{t}}_j}{\|\mathbf{f}^{\text{v}}_i\| \|\mathbf{f}^{\text{t}}_j\|}.
\end{equation}
These similarities are partitioned into two sets: $\mathcal{S}_{\text{gt}}$, the set of $s_{ij}$ values for ground-truth (GT) pairs, and $\mathcal{S}_{\text{non}}$, the set of all other non-GT pairs. We compute the mean ($\mu_x$) and standard deviation ($\sigma_x$) for both sets, where $x \in \{\text{gt}, \text{non}\}$:
\begin{equation}
    \mu_{x} = \frac{1}{|\mathcal{S}_{x}|} \sum_{s \in \mathcal{S}_{x}} s, \quad \sigma_{x} = \sqrt{\frac{1}{|\mathcal{S}_{x}|} \sum_{s \in \mathcal{S}_{x}} (s - \mu_{x})^2}.
\end{equation}

Based on these statistics, we define a valid sampling interval $[\theta_{\text{min}}, \theta_{\text{max}}]$. 
Empirically, GT-pair similarities exhibit a higher mean than non-pairs, but the ranges partially overlap, making fixed thresholds unreliable. 
To robustly separate trivial negatives from potential positives, we adopt percentile-based thresholds: the lower bound $\theta_{\text{min}}$ is set to the $\alpha$-th percentile of the non-pair distribution $\mathcal{S}_{\text{non}}$, filtering out dissimilar clips that provide little meaningful training signal, while the upper bound $\theta_{\text{max}}$ is set to the $\beta$-th percentile of the GT-pair distribution $\mathcal{S}_{\text{gt}}$, removing highly similar clips that may act as false negatives:
\begin{align}
    \theta_{\text{min}} &= \text{Percentile}_{\alpha}(\mathcal{S}_{\text{non}}), \\
    \theta_{\text{max}} &= \text{Percentile}_{\beta}(\mathcal{S}_{\text{gt}}).
\end{align}



During training, given a target video $\mathbf{V}_A$ and its query $\mathbf{Q}_j$ (indexed by $j$), we randomly sample another video $\mathbf{V}_B = \{\mathbf{c}_k\}_{k=1}^L$ from the dataset to act as the replacement source. We then construct an instance-specific candidate pool, $\mathcal{C}_{\text{pool}}$, containing only the clips from $\mathbf{V}_B$ whose similarity $s_{kj}$ to the query $\mathbf{Q}_j$ falls within the valid range (i.e., $\mathcal{C}_{\text{pool}} = \{\mathbf{c}_k \in \mathbf{V}_B \mid \theta_{\text{min}} \le s_{kj} \le \theta_{\text{max}}\}$). When generating $\mathbf{V}_{\text{mix}}$, the regions in $\mathbf{V}_A$ where $\mathbf{M}=0$ are filled by randomly sampling clips only from this filtered $\mathcal{C}_{\text{pool}}$.

Furthermore, we employ a context-preserving strategy. Given the set of GT indices $\mathcal{G}$, we define an extended context window $\mathcal{G}_{\text{ext}}$ by including $p$ adjacent clips immediately before the starting boundary and $p$ adjacent clips immediately after the ending boundary of the GT segment. The preserving mask $\mathbf{M}$ is set to $1$ for all indices $i \in \mathcal{G}_{\text{ext}}$, ensuring the target moment and its crucial surrounding context are not corrupted.
This two-pronged approach ensures a robust context diversification: it avoids false negatives by sampling only semantically unrelated yet non-trivial clips, while simultaneously preserving the essential temporal context required for precise moment reasoning.

\subsection{Context-enhanced Transformer Encoder}

Most existing moment retrieval models often perform immediate cross-attention between video features and the text query, without adequately modeling the video's internal temporal context. However, to distinguish fine-grained semantic moments, capturing the relationships between adjacent clips is crucial. To effectively capture this local-to-global temporal context, we propose the \textit{Context-enhanced Transformer Encoder (CTE)}.

The CTE consists of $N_{\text{b}}$ stacked blocks. Let the inputs to the $l$-th block be the video features $\mathbf{F}^{(l-1)} \in \mathbb{R}^{L \times D_{\text{v}}}$ and the learnable query features $\mathbf{Q}^{(l-1)} \in \mathbb{R}^{M \times D_{\text{v}}}$. The initial inputs are $\mathbf{F}^{(0)} = \mathbf{F}_{\text{v}}$ and a set of initialized learnable queries $\mathbf{Q}^{(0)}$. Each block consists of self-attention layers followed by a bidirectional cross-attention layer.

First, we process the video and query features independently. For the video features $\mathbf{F}^{(l-1)}$, inspired by \cite{liu2021swin}, we partition them into $K = L/W$ non-overlapping windows, where each window forms a windowed feature $\{\bar{\mathbf{F}}^{(l-1)}_k\}_{k=1}^K$. We then apply self-attention to these windowed features and reassemble them into the original sequence length to capture local temporal patterns:
\begin{equation}
\begin{gathered}
    \mathbf{F}' = \text{Concat}_{k=1}^K (\text{SelfAttn}(\bar{\mathbf{F}}^{(l-1)}_k)).
\end{gathered}
\end{equation}
We apply standard global self-attention to the learnable queries $\mathbf{Q}^{(l-1)}$:
\begin{equation}
    \mathbf{Q}' = \text{SelfAttn}(\mathbf{Q}^{(l-1)}).
\end{equation}

Next, to facilitate information exchange between the local video contexts and the global query representations, we perform bidirectional cross-attention. The video features are refined by attending to the queries, and the queries are refined by attending to the video features:
\begin{equation}
\begin{aligned}
    \hat{\mathbf{F}} &= \text{CrossAttn}(\mathbf{F}', \mathbf{Q}', \mathbf{Q}'), \\
    \hat{\mathbf{Q}} &= \text{CrossAttn}(\mathbf{Q}', \mathbf{F}', \mathbf{F}'),
\end{aligned}
\end{equation}
which are then passed through separate Feed-Forward Networks (FFN) with residual connections to produce the block outputs:
\begin{equation}
\begin{aligned}
    \mathbf{F}^{(l)} &= \text{Norm}(\text{MLP}(\hat{\mathbf{F}})) + \mathbf{F}^{(l-1)}, \\
    \mathbf{Q}^{(l)} &= \text{Norm}(\text{MLP}(\hat{\mathbf{Q}})) + \mathbf{Q}^{(l-1)}.
\end{aligned}
\end{equation}
These outputs serve as the inputs for the $(l+1)$-th block.

After $N_{\text{b}}$ layers, we aggregate the hierarchical contexts by concatenating the video outputs from all blocks, $\mathbf{F}_{\text{b}} = \text{Concat}_{l=1}^{N_{\text{b}}} (\mathbf{F}^{(l)})$. This aggregated feature is processed and combined with the original input features via a learnable weighted sum parameterized by $\omega$, producing the final context-enhanced features $\mathbf{F}_{\text{CTE}}$:
\begin{equation}
    \mathbf{F}_{\text{CTE}} = \omega \cdot \mathbf{F}_{\text{v}} + (1-\omega) \cdot \text{Norm}(\text{MLP}(\mathbf{F}_{\text{b}})).
\end{equation}

Subsequently, $\mathbf{F}_{\text{CTE}}$ and the text query features $\mathbf{F}_{\text{t}}$ are fed into a multimodal encoder. This encoder consists of a sequence of cross-attention, self-attention, 1D convolution, and self-attention layers to produce the final query-aligned multimodal features, $\mathbf{F}_{\text{m}}$. These features $\mathbf{F}_{\text{m}}$ are then utilized by two downstream components: (1) a Transformer Decoder that predicts the temporal spans and saliency scores, and (2) a Contrastive Head (composed of an MLP) which projects the features for our Context-invariant Boundary Discrimination (CBD) loss, detailed in \secref{sec:CBD}.

\subsection{Context-invariant Boundary Discrimination}
\label{sec:CBD}

Our QCD strategy generates augmented videos where the temporal context is altered, while the semantics of the target moment remain the same. The features within the target moment must remain semantically consistent across those videos. To enforce this, we propose the \textit{Context-invariant Boundary Discrimination (CBD) loss}, which learns this consistency by focusing on the regions where alignment is most challenging: the temporal boundaries.

The CBD loss is designed to focus explicitly on the features at the moment's boundaries. Given two augmentations, $\mathbf{V}'_{\text{mix}}$ and $\mathbf{V}''_{\text{mix}}$, generated from the same video, we first obtain their multimodal features $\mathbf{F}'_{\text{m}} = \{\mathbf{f}'_{{\text{m}},i}\}_{i=1}^L$ and $\mathbf{F}''_{\text{m}} = \{\mathbf{f}''_{{\text{m}},i}\}_{i=1}^L$ from the multimodal encoder. For simplicity, we describe the process for a single ground-truth (GT) span, denoted by the index set $\mathcal{G}$; this is applied to all GT spans present. We define the set of boundary indices as $\mathcal{B} = \{\min(\mathcal{G}),\max(\mathcal{G})\}$.

For each boundary index $b \in \mathcal{B}$, we construct the anchor set $\mathcal{Z}$ by collecting the features at index $b$ from the first augmentation $\mathbf{F}'_{\text{m}}$. The corresponding positive set $\mathcal{Z}^{+}$ is formed by the feature at the same temporal indices from the second augmentation $\mathbf{F}''_{\text{m}}$:
\begin{equation}
\begin{gathered}
    \mathcal{Z} = \{\mathbf{z}_{b}\}_{b\in \mathcal{B}},~~\mathbf{z}_{b} = \text{MLP} (\mathbf{f}'_{{\text{m}}, b}),\\ 
    \mathcal{Z}^+ = \{\mathbf{z}^+_{b}\}_{b\in \mathcal{B}},~~\mathbf{z}_{b}^{+} = \text{MLP} (\mathbf{f}''_{{\text{m}}, b}).
\end{gathered}
\end{equation}

We then construct the set of hard negative samples $\mathcal{Z}^-$ for these anchors from two distinct sources, targeting both temporal and semantic ambiguity. First, we define the set of all background indices $\mathcal{I}_{\text{bg}} = \{1, \dots, L\} \setminus \mathcal{G}$. To ensure precise boundary discrimination, we identify the set of temporally adjacent background indices $\mathcal{I}^{\text{adj}}_b$ relative to the anchor's index $b$:
\begin{equation}
    \mathcal{I}^{\text{adj}}_b = \{ j \mid j \in \mathcal{I}_{\text{bg}} \text{ and } |j - b| \le N_{\text{adj}} \}.
\end{equation}

Second, to account for semantically confusable clips that are temporally distant, we mine the $N_\text{hard}$ hardest negatives from the remaining background $\mathcal{I}_b^{\text{rem}} = \mathcal{I}_{\text{bg}} \setminus \mathcal{I}_b^{\text{adj}}$. These indices, $\mathcal{I}_b^{\text{hard}}$, are selected based on the highest cosine similarity $s(\cdot, \cdot)$ between the boundary feature and the background features $\mathbf{f}'_{m,j}$:
\begin{equation}
    \mathcal{I}_b^{\text{hard}} = \underset{j \in \mathcal{I}_b^{\text{rem}}}{\text{top-}N_{\text{hard}}} \text{ } s(\mathbf{f}'_{m, b}, \mathbf{f}'_{m, j}).
\end{equation}

The final set of negative samples $\mathcal{Z}^-$ is the union of the projected features through the contrastive head corresponding to these two index sets:
\begin{equation}
    \mathcal{Z}^- = \{ \text{MLP}(\mathbf{f}'_{m, j}) \mid j \in \mathcal{I}_b^{\text{adj}} \cup \mathcal{I}_b^{\text{hard}}, {b\in \mathcal{B}} \}.
\end{equation}

The final CBD loss is the average of the contrastive losses computed for each boundary anchor $b \in \mathcal{B}$. Each anchor $\mathbf{z}_b$ is contrasted against its corresponding positive $\mathbf{z}^+_b$ and the entire set of negative samples $\mathcal{Z}^-$:
\begin{equation}
\begin{aligned}
    \mathcal{L}_{CBD} &= - \frac{1}{|\mathcal{B}|} \sum_{b \in \mathcal{B}} \log \frac{\exp(s_{p,b})}{\exp(s_{p,b}) + \sum_{\mathbf{z}_{n} \in \mathcal{Z}^-}\exp(s_{n,b})}, \\
    s_{p,b} &= s(\mathbf{z}_{b}, \mathbf{z}^{+}_{b}) / \tau,~~
     s_{n,b} = s(\mathbf{z}_{b}, \mathbf{z}_{n}) / \tau.
\end{aligned}
\end{equation}
Here, $\tau$ is a temperature hyper-parameter. By explicitly enforcing invariance on the boundary features against this comprehensive set of hard negatives, $\mathcal{L}_{\text{CBD}}$ guides the model to learn a highly discriminative representation, enhancing the precision of moment localization.

\begin{table*}[!t]
\caption{\textbf{MR and HD results on the QVHighlights test split.} Our model demonstrates notably higher performance in Rank-1 (R1) and Highlight Detection (HD), indicating its superior ability to capture fine-grained moment boundaries.}
\begin{adjustbox}{max width=\textwidth, center}
\begin{tabular}{c c c c c c c c}
\toprule
\multirow{3}{*}{\textbf{Method}} & \multicolumn{5}{c}{\textbf{Moment Retrieval}} & \multicolumn{2}{c}{\textbf{Highlight Detection}} \\ \cline{2-6} \cline{7-8}
& \multicolumn{2}{c}{R1} & \multicolumn{3}{c}{mAP} & \multicolumn{2}{c}{$\ge$Very Good} \\ \cline{2-3} \cline{4-6} \cline{7-8}
& @0.5 & @0.7 & @0.5 & @0.75 & Avg. & mAP & HIT@1 \\ \hline
\multicolumn{1}{l}{MCN \cite{anne2017localizing} \conf{[ICCV'17]}} & 11.41 & 2.72 & 24.94 & 8.22 & 10.67 & – & – \\
\multicolumn{1}{l}{CAL \cite{escorcia2019finding} \conf{[arXiv'19]}} & 25.49 & 11.54 & 23.40 & 7.65 & 9.89 & – & – \\
\multicolumn{1}{l}{XML \cite{lei2020tvr} \conf{[ECCV'20]}} & 41.83 & 30.35 & 44.63 & 31.73 & 32.14 & 34.49 & 55.25 \\
\multicolumn{1}{l}{Moment-DETR \cite{lei2021Qvhighlights} \conf{[NeurIPS'21]}} & 52.89 & 33.02 & 54.82 & 29.40 & 30.73 & 35.69 & 55.60 \\
\multicolumn{1}{l}{UMT \cite{liu2022umt} \conf{[CVPR'22]}} & 56.23 & 41.18 & 53.83 & 37.01 & 36.12 & 38.18 & 59.99 \\
\multicolumn{1}{l}{MomentDiff \cite{li2023momentdiff} \conf{[NeurIPS'24]}} & 57.42 & 39.66 & 54.02 & 35.73 & 35.95 & – & – \\
\multicolumn{1}{l}{UniVTG \cite{lin2023univtg} \conf{[ICCV'23]}} & 58.86 & 40.86 & 57.60 & 35.59 & 35.47 & 38.20 & 60.96 \\
\multicolumn{1}{l}{BM-DETR \cite{jung2025background} \conf{[WACV'25]}} & 60.12 & 43.05 & 63.08 & 40.18 & 40.08 & – & – \\
\multicolumn{1}{l}{EaTR \cite{jang2023knowing} \conf{[ICCV'23]}} & 61.36 & 45.79 & 61.86 & 41.91 & 41.74 & 37.15 & 58.65 \\
\multicolumn{1}{l}{QD-DETR \cite{moon2023query} \conf{[CVPR'23]}} & 62.40 & 44.98 & 62.52 & 39.88 & 39.86 & 38.94 & 62.40 \\
\multicolumn{1}{l}{BAM-DETR \cite{lee2024bam} \conf{[ECCV'24]}} & 62.71 & 48.64 & 64.57 & 46.33 & 45.36 & – & – \\
\multicolumn{1}{l}{MESM \cite{liu2024towards} \conf{[AAAI'24]}} & 62.78 & 45.20 & 62.64 & 41.45 & 40.68 & – & – \\
\multicolumn{1}{l}{UVCOM \cite{xiao2024bridging} \conf{[CVPR'24]}} & 63.55 & 47.47 & 63.37 & 42.67 & 43.18 & 39.74 & 64.20 \\
\multicolumn{1}{l}{TD-DETR \cite{zhou2025devil} \conf{[ICCV'25]}} & 64.53 & \underline{50.37} & 66.21 & \underline{47.32} & \underline{46.69} & – & – \\
\multicolumn{1}{l}{TR-DETR \cite{sun2024tr} \conf{[AAAI'24]}} & 64.66 & 48.96 & 63.98 & 43.73 & 42.62 & 39.91 & 63.42 \\
\multicolumn{1}{l}{CG-DETR \cite{moon2023correlation} \conf{[arXiv'23]}} & 65.43 & 48.38 & 64.51 & 42.77 & 42.86 & \underline{40.33} & \textbf{66.21} \\
\multicolumn{1}{l}{CDTR \cite{ran2025cdtr} \conf{[AAAI'25]}} & \underline{65.79} & 49.60 & \underline{66.44} & 45.96 & 44.37 & – & – \\ 
\rowcolor{gray!15} \multicolumn{1}{l}{\textbf{Ours}} & \textbf{70.05} & \textbf{55.32} & \textbf{69.49} & \textbf{48.45} & \textbf{47.49} & \textbf{44.43} & \underline{66.01} \\ \bottomrule
\end{tabular}
\end{adjustbox}
\label{tab:qvh}
\end{table*}

Our final training objective jointly optimizes the Moment Retrieval (MR) loss $\mathcal{L}_{\text{MR}}$, the Highlight Detection (HD) loss $\mathcal{L}_{\text{HD}}$, and our proposed Context-invariant Boundary Discrimination (CBD) loss $\mathcal{L}_{\text{CBD}}$. Following \cite{lei2021Qvhighlights, carion2020end}, we first perform an optimal bipartite matching between the predicted moments and the ground-truth moments to assign pairs for loss calculation. The overall objective $\mathcal{L}_{\text{total}}$ is formulated as:
\begin{equation}
\begin{aligned}
    \mathcal{L}_{\text{MR}} &= \lambda_{\text{L1}} \mathcal{L}_{\text{L1}} + \lambda_{\text{gIoU}} \mathcal{L}_{\text{gIoU}} \\
    \mathcal{L}_{\text{HD}} &= \lambda_{\text{HD}}(\mathcal{L}_{\text{margin}} + \mathcal{L}_{\text{rank}}) \\
    \mathcal{L}_{\text{total}} &= \mathcal{L}_{\text{MR}} + \mathcal{L}_{\text{HD}} + \lambda_{\text{CBD}} \mathcal{L}_{\text{CBD}},
\end{aligned}
\end{equation}
where $\mathcal{L}_{\text{L1}}$ and $\mathcal{L}_{\text{gIoU}}$ are the L1 loss and GIoU loss \cite{rezatofighi2019generalized} for MR, and $\mathcal{L}_{\text{margin}}$ and $\mathcal{L}_{\text{rank}}$ are the margin ranking and rank-aware losses for HD \cite{lei2021Qvhighlights}. $\lambda_{\text{L1}}, \lambda_{\text{gIoU}}, \lambda_{\text{HD}}, \text{and } \lambda_{\text{CBD}}$ are hyperparameters to balance the different loss components.

\section{Experiments}
\subsection{Experimental Setup}
\noindent{\textbf{Datasets}}
We evaluate our method on three public benchmarks: QVHighlights, Charades-STA, and TACoS.

\begin{itemize}
    \item \textbf{QVHighlights} \cite{lei2021Qvhighlights} is a large-scale dataset containing 10,148 YouTube videos. Each video is paired with at least one text query that annotates a specific highlight moment. Following prior work, we report results using the official CodaLab evaluation server.

    \item \textbf{Charades-STA} \cite{sigurdsson2016charades} is derived from the Charades dataset and focuses on daily indoor activities. It consists of 9,848 videos and 16,128 human-annotated query descriptions. We adhere to the standard protocol, using 12,408 video-query pairs for training and 3,720 for testing.

    \item \textbf{TACoS} \cite{regneri2013tacos} features 127 long videos centered around cooking activities. This dataset provides a challenging scenario due to the extended video durations and complex temporal dynamics.
\end{itemize}

\noindent{\textbf{Evaluation Metrics}}~
We evaluate our model using three standard metrics. \textbf{Recall@1 (R@1)} measures the percentage of top-1 predictions whose IoU exceeds a given threshold, reported at $\{0.5, 0.7\}$. \textbf{Mean Average Precision (mAP)} computes the average precision across various IoU thresholds, and \textbf{Mean IoU (mIoU)} is the average IoU over all test samples.

\noindent{\textbf{Implementation details}}~
Following prior work \cite{lei2021Qvhighlights}, we utilize video features extracted from both a pre-trained SlowFast model \cite{feichtenhofer2019slowfast} and a CLIP 
vision encoder \cite{radford2021learning}. Text features are extracted using the 
corresponding CLIP text encoder. The network parameters are optimized using the AdamW 
optimizer \cite{loshchilov2017decoupled} with a cosine annealing learning rate scheduler.
For the QVHighlights dataset, we train for 250 epochs with a batch size of 32. 
For Charades-STA and TACoS, we train for 100 epochs with a batch size of 32. 
For all datasets, the initial learning rate is set to $1 \times 10^{-4}$ with 
a weight decay of $1 \times 10^{-4}$. The coefficients for losses are set to $\lambda_{\text{L1}}=10$, $\lambda_{\text{gIoU}}=1$, 
$\lambda_{\text{HD}}=1$, and $\lambda_{\text{CBD}}=0.005$ by default. For QCD, the percentile thresholds are set to $\alpha{=}10$ and $\beta{=}60$ with a replacement ratio of 0.3 and context preservation window $p{=}1$ across all datasets.
These QCD settings are fixed and used as the default baseline for CTE and CBD ablations.

\subsection{Comparison to State-of-the-art}
\noindent{\textbf{QVHighlights}~
We compare our CVA model to state-of-the-art methods on the QVHighlights test split, as shown in Tab.~\ref{tab:qvh}. CVA outperforms all competing approaches, achieving substantial improvements across the majority of evaluation metrics. The improvements are most pronounced in Moment Retrieval (MR) recall, highlighting the effectiveness of our QCD strategy in mitigating false negatives. Specifically, CVA achieves a substantial gain of 4.95 on R1@0.7 over the previous best method, TD-DETR \cite{zhou2025devil}, and 4.26 on R1@0.5 over CDTR \cite{ran2025cdtr}. 
Furthermore, our Context-invariant Boundary Discrimination (CBD) strategy leads to more precise saliency predictions, resulting in a new state-of-the-art for Highlight Detection (HD) mAP. Our score of 44.43 surpasses the previous best from CG-DETR \cite{moon2023correlation} by 4.1 points. CVA also achieves the highest MR Average mAP at 47.49 and a highly competitive HIT@1 score, demonstrating the synergistic benefits of our context-aware framework.

\noindent{\textbf{TACoS}}~
We further validate our CVA on the challenging TACoS dataset, with results presented in \tabref{tab:tacos_comparison}. Our method consistently outperforms previous state-of-the-art approaches, establishing new performance benchmarks across all reported metrics. Notably, CVA achieves a mIoU of 41.07, surpassing the previous best, BAM-DETR \cite{lee2024bam}, by 1.76. Similar improvements are observed across recall metrics, including R1@0.5 (43.21 vs. 41.45) and R1@0.7 (27.73 vs. 26.77). This strong performance on TACoS, a benchmark known for its long videos and complex temporal dependencies, underscores the effectiveness of our context-aware framework in precisely grounding queries within extended temporal contexts.

\noindent{\textbf{Charades-STA}}~
We also evaluate our CVA on the Charades-STA benchmark, with results shown in Tab.~\ref{tab:charades_sta_comparison}. CVA achieves the new state-of-the-art performance across all metrics, surpassing prior methods by a clear margin. In particular, CVA attains \textbf{62.61} R1@0.5 and \textbf{40.78} R1@0.7, outperforming the previous best results from BAM-DETR \cite{lee2024bam} by \textbf{+2.66} and \textbf{+1.40}, respectively.
Our method also achieves the highest mIoU of \textbf{53.35}, improving over BAM-DETR by \textbf{+1.02}.
These results confirm that our context-aware design generalizes effectively beyond QVHighlights, providing superior temporal grounding capability even in diverse indoor activity scenarios such as Charades-STA.

\begin{table}[t!]
    \centering
    \caption{Comparison on the TACoS test split.}
    \label{tab:tacos_comparison}
    \begin{adjustbox}{max width=0.45\textwidth}
    \begin{tabular}{lcccc}
        \toprule
        \multirow{2}{*}{Method} & \multicolumn{3}{c}{R1} & \multirow{2}{*}{mIoU} \\
        \cmidrule(lr){2-4}
         & @0.3 & @0.5 & @0.7 &  \\
        \midrule
        2D-TAN \cite{zhang2020learning} \conf{[AAAI'20]} & 40.01 & 27.99 & 12.92 & 27.22 \\
        VSLNet \cite{zhang2020span} \conf{[ACL'20]} & 35.54 & 23.54 & 13.15 & 24.99 \\
        M-DETR \cite{lei2021Qvhighlights} \conf{[NeurIPS'21]} & 37.97 & 24.67 & 11.97 & 25.49 \\
        MomentDiff \cite{li2023momentdiff} \conf{[NeurIPS'23]} & 44.78 & 33.68 & - & - \\
        UniVTG \cite{lin2023univtg} \conf{[ICCV'23]} & 51.44 & 34.97 & 17.35 & 33.60 \\
        UVCOM \cite{xiao2024bridging} \conf{[CVPR'24]} & - & 36.39 & 23.32 & - \\
        CDTR \cite{ran2025cdtr} \conf{[AAAI'25]} & 53.41 & 40.26 & 23.43 & 37.28 \\
        BAM-DETR \cite{lee2024bam} \conf{[ECCV'24]} & 56.69 & 41.45 & 26.77 & 39.31 \\
        \rowcolor{gray!15} \textbf{Ours} & \textbf{58.80} & \textbf{43.21} & \textbf{27.73} & \textbf{41.07} \\
        \bottomrule
    \end{tabular}
    \end{adjustbox}
\end{table}

\begin{table}[t]
    \centering
    \caption{Comparison on the Charades-STA valid split.}
    \label{tab:charades_sta_comparison}
    \begin{adjustbox}{max width=0.45\textwidth}
    \begin{tabular}{lcccc}
        \toprule
        \multirow{2}{*}{Method} & \multicolumn{3}{c}{R1} & \multirow{2}{*}{mIoU} \\
        \cmidrule(lr){2-4}
         & @0.3 & @0.5 & @0.7 &  \\
        \midrule
        2D-TAN \cite{zhang2020learning} \conf{[AAAI'20]} & 58.76 & 46.02 & 27.50 & 41.25 \\
        VSLNet \cite{zhang2020span} \conf{[ACL'20]} & 60.30 & 42.69 & 24.14 & 41.58 \\
        M-DETR \cite{lei2021Qvhighlights} \conf{[NeurIPS'21]} & 65.83 & 52.07 & 30.59 & 45.54 \\
        MomentDiff \cite{li2023momentdiff} \conf{[NeurIPS'23]} & - & 55.57 & 32.42 & - \\
        QD-DETR \cite{moon2023query} \conf{[CVPR'23]} & - & 57.31 & 32.55 & - \\
        UniVTG \cite{lin2023univtg} \conf{[ICCV'23]} & 70.81 & 58.01 & 35.65 & 50.10 \\
        TR-DETR \cite{sun2024tr} \conf{[AAAI'24]} & - & 57.61 & 33.52 & - \\
        UVCOM \cite{xiao2024bridging} \conf{[CVPR'24]}  & - & 59.25 & 36.64 & - \\
        BAM-DETR \cite{lee2024bam} \conf{[ECCV'24]} & 72.93 & 59.95 & 39.38 & 52.33 \\
        CDTR \cite{ran2025cdtr} \conf{[AAAI'25]} & 71.16 & 60.39 & 37.24 & 50.65 \\
        \rowcolor{gray!15} \textbf{Ours} & \textbf{74.19} & \textbf{62.61} & \textbf{40.78} & \textbf{53.35} \\
        \bottomrule
    \end{tabular}
    \end{adjustbox}
\end{table}

\subsection{Ablation Study}

\noindent{\textbf{Main Component Analysis}}~
All ablation studies are conducted on the QVHighlights valid split. We conduct an ablation study to validate the effectiveness of our three key components: the Query-aware Context Diversification (QCD), the Context-enhanced Transformer Encoder (CTE), and Context-invariant Boundary Discrimination (CBD). The results are presented in \tabref{tab:ablation_cte_qcd_cbd}. 
Our baseline model (Row 1) achieves an R1@0.7 of 46.77 and an HD mAP of 37.80. 
Applying the QCD augmentation alone (Row 2) yields significant improvements (+5.21 in R1@0.7 and +3.92 in HD mAP), confirming the effectiveness of our query-aware sampling strategy. 
Building upon QCD, adding CTE (Row 3) further improves performance (e.g., 52.63 R1@0.7), demonstrating the benefit of hierarchical temporal context modeling. 
Similarly, adding CBD on top of QCD (Row 4) brings notable gains in both MR and HD metrics (e.g., 53.02 R1@0.7 and 42.87 HD mAP), validating the effectiveness of boundary-focused contrastive learning.
Finally, our full model (Row 5), which integrates all three components, achieves the best performance across all metrics (e.g., 54.84 in R1@0.7 and 43.47 in HD mAP). This clearly validates that QCD, CTE, and CBD are all effective and contribute synergistically to our model's final performance.

\begin{table}[t]
\centering
\caption{Ablation study on our main components: Query-aware Context Diversification (QCD), Context-enhanced Transformer Encoder (CTE), and Context-invariant Boundary Discrimination (CBD).}
\label{tab:ablation_cte_qcd_cbd}
\begin{adjustbox}{max width=\linewidth}
\begin{tabular}{c c c|ccc|cc}
\toprule
\multirow{2}{*}{\textbf{QCD}} & 
\multirow{2}{*}{\textbf{CTE}} & 
\multirow{2}{*}{\textbf{CBD}} & 
\multicolumn{3}{c|}{\textbf{Moment Retrieval}} & 
\multicolumn{2}{c}{\textbf{Highlight Detection}} \\ \cline{4-8}
 &  &  & R1@0.5 & R1@0.7 & \multicolumn{1}{c|}{mAP@0.5} & mAP & HIT@1 \\ 
\midrule
 &  &  & 61.12 & 46.77 & \multicolumn{1}{c|}{62.45} & 37.80 & 61.47 \\
\checkmark &  &  & 65.32 & 51.98 & \multicolumn{1}{c|}{65.92} & 41.72 & \underline{68.44} \\
\checkmark & \checkmark &  & 67.62 & 52.63 & \multicolumn{1}{c|}{\underline{67.81}} & 41.89 & 68.26 \\
\checkmark &  & \checkmark & \underline{67.87} & \underline{53.02} & \multicolumn{1}{c|}{66.89} & \underline{42.87} & 68.19 \\
\checkmark & \checkmark & \checkmark & \textbf{69.61} & \textbf{54.84} & \multicolumn{1}{c|}{\textbf{67.97}} & \textbf{43.47} & \textbf{70.40} \\
\bottomrule
\end{tabular}
\end{adjustbox}
\end{table}
\begin{table}[t]
\centering
\caption{Ablation study on Query-aware
Context Diversification (QCD). (a) Comparison of data augmentation strategies, (b) Sensitivity to percentile thresholds $(\alpha, \beta)$.}
\label{tab:qcd_ablation_overall}

\begin{subtable}{\linewidth}
\centering
\caption{Query-aware vs Query-agnostic augmentation}
\label{tab:qcd_results_sub}
\begin{adjustbox}{max width=\linewidth}
\begin{tabular}{l|ccc|cc}
\toprule
Setting & R1@0.5 & R1@0.7 & mAP@0.5 & mAP & HIT@1 \\ 
\midrule
Baseline \cite{zhou2025devil} & 61.12 & 46.77 & 62.45 & 37.80 & 61.47 \\
+ Query-agnostic mix & 63.08 & 48.39 & 63.59 & 39.10 & 63.68 \\
\rowcolor{gray!15}
+ QCD (Ours) & \textbf{65.32} & \textbf{51.98} & \textbf{65.92} & \textbf{41.72} & \textbf{68.44} \\
\bottomrule
\end{tabular}
\end{adjustbox}
\end{subtable}

\vspace{5mm}

\begin{subtable}{\linewidth}
\centering
\caption{Percentile Threshold Sensitivity}
\label{tab:min_max_table_sub}
\begin{adjustbox}{max width=\linewidth}
\begin{tabular}{cc|ccc|cc}
\toprule
\textbf{$\alpha$ (non-pair)} & \textbf{$\beta$ (GT-pair)} &
\multicolumn{3}{c|}{MR} & \multicolumn{2}{c}{HD} \\ \cline{3-7}
 &  & R1@0.5 & R1@0.7 & mAP@0.5 & mAP & HIT@1 \\ 
\midrule
10 & 90 & 63.64 & 48.88 & 64.71 & 38.66 & 65.43 \\
40 & 90 & 63.20 & 49.65 & 64.12 & 39.46 & 66.30 \\
40 & 60 & \textbf{64.68} & \underline{49.71} & \underline{64.76} & \textbf{41.71} & \underline{67.65} \\
\rowcolor{gray!15}
10 & 60 & \underline{64.45} & \textbf{50.92} &
\textbf{65.37} & \underline{41.38} & \textbf{68.02} \\
\bottomrule
\end{tabular}
\end{adjustbox}
\end{subtable}

\end{table}

\noindent{\textbf{Analysis of QCD Components}}~
We first compare our Query-aware Context Diversification (QCD) with the query-agnostic mix
strategy from \cite{zhou2025devil}. As shown in Tab.~\ref{tab:qcd_ablation_overall}-(a), the
query-agnostic mix yields only marginal improvements over the baseline model, while QCD provides
substantially larger gains across all metrics (e.g., \textbf{+5.21 R1@0.7} and \textbf{+3.92 HD mAP}). 
These results highlight that conditioning augmentation on the query is crucial for preventing 
false-negative contamination and for improving both moment retrieval and highlight discrimination.

Tab.~\ref{tab:qcd_ablation_overall}-(b) further analyzes QCD by varying the percentile 
thresholds $(\alpha, \beta)$ that determine the valid similarity range for sampling replacement clips. Setting a high upper bound (e.g., $\beta=90$) retains many GT-like clips in the candidate 
pool, which increases the risk of false-negative contamination and leads to limited improvements. 
On the other hand, while setting a high lower bound (e.g., $\alpha=40$) successfully removes more trivial negatives, it overly restricts the valid sampling range, significantly reducing the overall diversity of the augmented contexts. 
Therefore, moderately restricting the upper bound ($\beta=60$) to prevent false negatives, while keeping the lower bound small ($\alpha=10$) to filter out only the most meaningless backgrounds while preserving augmentation diversity, yields the best balance and the highest performance (e.g., \textbf{50.92 R1@0.7} and \textbf{68.02 HIT@1}).
All subsequent component ablations (CTE, CBD) are conducted with QCD applied as the default augmentation, as reflected in \tabref{tab:ablation_cte_qcd_cbd}.

\begin{table}[t]
\centering
\caption{Ablation results for our CBD loss. We analyze the number of negative samples taken from the adjacent temporal margin ($N_{\text{adj}}$) and the $N_{\text{hard}}$ semantically hardest negative samples. The baseline row (-, -) indicates the performance with only the QCD augmentation applied.}
\label{tab:CBD_results}
\begin{adjustbox}{max width=\linewidth}
\begin{tabular}{cc|ccc|cc}
\toprule
\multirow{2}{*}{\textbf{$N_{\text{adj}}$}} &
\multirow{2}{*}{\textbf{$N_{\text{hard}}$}} &
\multicolumn{3}{c|}{\textbf{Moment Retrieval}} &
\multicolumn{2}{c}{\textbf{Highlight Detection}} \\ 
 &  & R1@0.5 & R1@0.7 & \multicolumn{1}{c|}{mAP@0.5} & mAP & HIT@1 \\ 
\midrule
- & - & 65.32 & 51.98 & \multicolumn{1}{c|}{65.92} & 41.72 & 68.44 \\ 
\midrule
2 & 0 & 67.89 & 51.94 & \multicolumn{1}{c|}{67.12} & 41.82 & 67.42 \\
3 & 0 & 68.12 & 51.98 & \multicolumn{1}{c|}{66.76} & 42.01 & 66.88 \\
2 & 2 & \underline{68.98} & \underline{54.89} & \multicolumn{1}{c|}{\textbf{68.02}} & 42.79 & 68.99 \\
\rowcolor{gray!15}
2 & 5 & \textbf{69.61} & 54.84 & \multicolumn{1}{c|}{\underline{67.97}} & \textbf{43.47} & \textbf{70.40} \\
2 & 7 & 68.76 & \textbf{54.97} & \multicolumn{1}{c|}{66.88} & \underline{43.03} & \underline{69.11} \\
2 & 10 & 67.76 & 52.71 & \multicolumn{1}{c|}{66.47} & 42.34 & 68.46 \\
3 & 5 & 68.52 & 53.12 & \multicolumn{1}{c|}{67.13} & 42.33 & 68.94 \\
\bottomrule
\end{tabular}
\end{adjustbox}
\end{table}

\noindent{\textbf{Analysis of CBD Loss}}~
We validate our Context-invariant Boundary Discrimination (CBD) loss in Table~\ref{tab:CBD_results}. The baseline (Row 1), trained with only the QCD augmentation, achieves 51.98 on R1@0.7. Simply adding the CBD loss with only temporally adjacent negatives (i.e., $N_{\text{hard}}{=}0$) provides a notable improvement (Rows 2-3). However, the most significant performance gain comes from introducing semantically hard negatives ($N_{\text{hard}}{>}0$). For instance, setting $(N_{\text{adj}}{=}2, N_{\text{hard}}{=}2)$ (Row 4) boosts R1@0.7 by nearly 3 points over $(N_{\text{adj}}{=}2, N_{\text{hard}}{=}0)$ (Row 2), confirming that addressing both temporal and semantic ambiguity is crucial. Our final configuration of $(N_{\text{adj}}{=}2, N_{\text{hard}}{=}5)$ (Row 5) achieves the best overall performance, including the highest HD mAP (43.47). This demonstrates that our hybrid negative sampling strategy, combining a small temporal margin with the $N_{\text{hard}}$ most semantically confusable clips, is highly effective for learning discriminative boundary representations.

\noindent{\textbf{Analysis of CTE Components}}~
Tab.~\ref{tab:ablation_cte_2} examines the two key mechanisms within our CTE: learnable queries and windowed self-attention.
The baseline model without either component serves as the starting point, exhibiting limited temporal modeling capability. 
Introducing only learnable queries yields consistent improvements across MR metrics 
(e.g., \textbf{+1.22 R1@0.5} and \textbf{+0.51 mAP@0.5}), while adding only windowed self-attention also 
provides notable gains, particularly in MR mAP@0.5 (\textbf{+0.29}). 
This demonstrates that local windowed attention and global learnable queries are complementary, jointly enhancing the model's ability to encode multi-scale temporal context.

\begin{table}[t]
\centering
\caption{Ablation on components of CTE, with QCD applied as the default augmentation.}
\label{tab:ablation_cte_2}
\begin{adjustbox}{max width=\linewidth}
\begin{tabular}{cc|ccc|cc}
\toprule
\multirow{2}{*}{\begin{tabular}[c]{@{}c@{}}learnable\\ queries\end{tabular}} & 
\multirow{2}{*}{\begin{tabular}[c]{@{}c@{}}window\\ self-attn\end{tabular}} & 
\multicolumn{3}{c|}{\textbf{Moment Retrieval}} & 
\multicolumn{2}{c}{\textbf{Highlight Detection}} \\ \cline{3-7}
 &  & R1@0.5 & R1@0.7 & mAP@0.5 & mAP & HIT@1 \\
\midrule
 &  & 65.32 & 51.98 & 65.92 & 41.72 & \textbf{68.44} \\
\checkmark &  & \underline{66.54} & \underline{52.33} & \underline{66.43} & \underline{41.85} & 67.66 \\
 & \checkmark & 65.50 & 52.13 & 66.21 & 41.13 & 68.23 \\
\rowcolor{gray!15}
\checkmark & \checkmark & \textbf{67.62} & \textbf{52.63} & \textbf{67.81} & \textbf{41.89} & \underline{68.26} \\
\bottomrule
\end{tabular}
\end{adjustbox}
\vspace{-3mm}
\end{table}

\section{Conclusion}
In this paper, we proposed \textbf{CVA (Context-aware Video-text Alignment)}, a novel framework designed to achieve robust video-text alignment that is both sensitive to temporal dynamics and invariant to irrelevant background context. Our approach tackles this challenge through a synergistic combination of data-centric and architectural enhancements.
Our method is built on three key components. We first introduced \textit{Query-aware Context Diversification (QCD)}, an advanced data augmentation strategy that simulates diverse contexts while preventing the false negative problem of query-agnostic mixing by using a similarity-based candidate pool. Second, we proposed the \textit{Context-invariant Boundary Discrimination (CBD) loss}, a contrastive objective that enforces semantic consistency at the challenging temporal boundaries, making the representations robust to contextual shifts. Third, our \textit{Context-enhanced Transformer Encoder (CTE)} utilizes windowed self-attention and bidirectional cross-attention to effectively capture multi-scale temporal context.
Through the synergy of these components, our experiments demonstrate that CVA sets a new state-of-the-art on major VTG benchmarks, including QVHighlights and Charades-STA, validating the effectiveness of our context-aware approach.

\paragraph{Acknowledgements.}
This work was supported by LG AI STAR Talent Development Program for Leading Large-Scale Generative AI Models in the Physical AI Domain (RS-2025-25442149), Basic Science Research Program through the National Research Foundation of Korea (NRF) funded by the Ministry of Education (RS-2025-25420118), the InnoCORE program of the Ministry of Science and ICT (N10250156) and the Institute of Information \& Communications Technology Planning \& Evaluation (IITP) grant funded by the Korea government (MSIT) (No. RS-2025-02219277, AI Star Fellowship Support Project (DGIST)).

{
    \small
    \bibliographystyle{ieeenat_fullname}
    \bibliography{main}
}

\clearpage
\appendix
\section*{Appendix}
\renewcommand{\thesection}{A.\arabic{section}}
\renewcommand{\thefigure}{A\arabic{figure}}
\renewcommand{\thetable}{A\arabic{table}}
\setcounter{section}{0}
\setcounter{figure}{0}
\setcounter{table}{0}
\renewcommand{\theequation}{A\arabic{equation}}
\setcounter{equation}{0}

\section{Additional Details about Query-aware Context Diversification (QCD)}

\textbf{Query-aware Context Diversification (QCD)} aims to generate synthetic training samples that are both 
\textbf{semantically safe} and \textbf{sufficiently informative}.  
A key challenge is to avoid two failure modes in the video--text similarity space:  
(1) \textbf{low-similarity outliers}, which yield overly trivial negatives, and  
(2) \textbf{high-similarity false negatives}, where clips that are semantically close to the query are mistakenly treated as background.  
QCD therefore focuses on selecting an \emph{intermediate} similarity band that provides meaningful augmentation without harming alignment.

\begin{figure}[H]
\centering
\includegraphics[width=\linewidth]{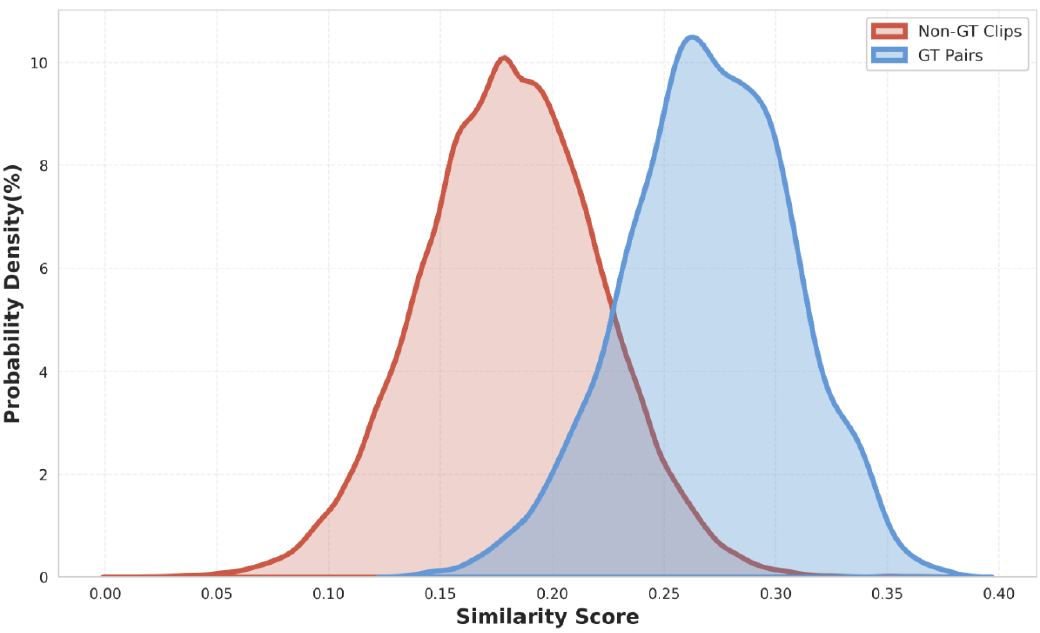}
\caption{Density distribution of cosine similarity across all clip--text pairs in QVHighlights dataset.  
The clustered structure motivates selecting an intermediate similarity band for QCD.}
\label{fig:overall}
\end{figure}

\subsection{Similarity Structure}

To characterize the global similarity structure, we compute cosine similarity between each text query and \emph{all} video clips in the QVHighlights dataset \cite{lei2021Qvhighlights}.  
This produces a dense set of clip--text similarity scores.  
Figure~\ref{fig:overall} visualizes the resulting distribution, where similarity values exhibit clear clustering rather than a uniform continuum.  
This landscape provides the basis for determining QCD’s safe and informative operating region.


\subsection{Avoiding Extremal Similarity Regions}

\paragraph{Low-similarity outliers.}  
Clips with very low similarity correspond to unrelated backgrounds.  
Replacing content with such clips produces synthetic samples that are too easy, offering limited benefit in learning robust temporal discrimination.

\paragraph{High-similarity false negatives.}  
Clips with high similarity often depict activities close to the queried action—even when originating from different videos.  
Using these clips as negative replacements introduces false-negative supervision, which can degrade retrieval accuracy.  
Examples in Fig.~\ref{fig:sim_example} illustrate such cases.

\begin{figure*}[t]
\centering
\includegraphics[width=0.85\textwidth]{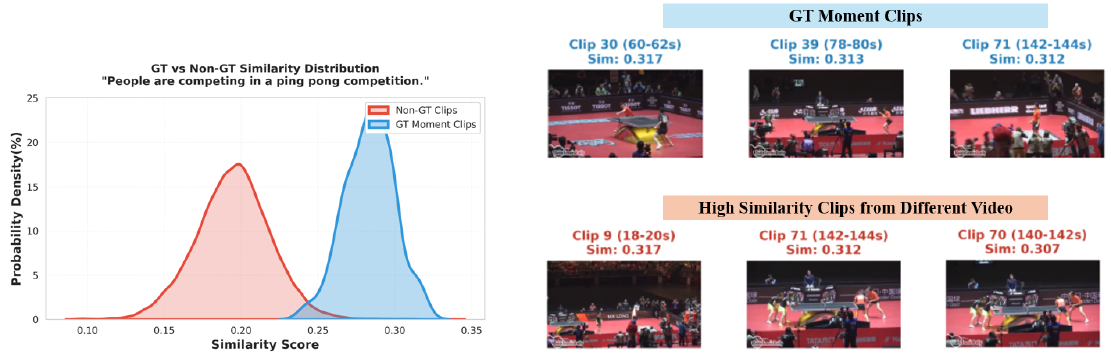}
\vspace{2mm}
\includegraphics[width=0.85\textwidth]{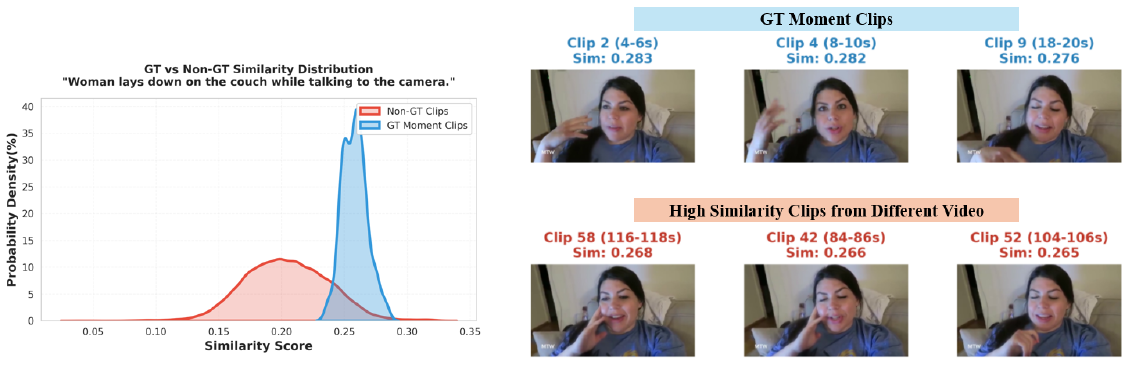}
\vspace{2mm}
\includegraphics[width=0.85\textwidth]{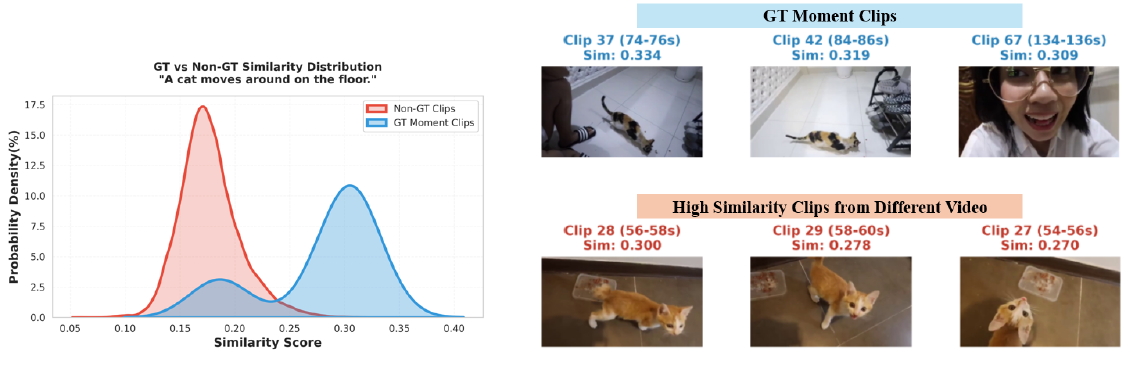}
\caption{Examples illustrating the risk of using high-similarity clips as negative replacements.  
Such clips contain semantically relevant actions, and including them introduces false-negative supervision.}
\label{fig:sim_example}
\end{figure*}

These observations support restricting QCD to an intermediate similarity band that excludes both extremes. In practice, we determine the intermediate similarity band 
$[\theta_{\min}, \theta_{\max}]$ based on percentile 
statistics of the global similarity distribution; the exact values are reported in the main paper.





\begin{table}[t]
\centering
\caption{Ablation on background replacement ratio and context preservation window size in QCD,
evaluated on the QVHighlights validation split.
\textbf{Replace Ratio} denotes the fraction of background clips marked as $M{=}0$ (non-MR region), and 
\textbf{Context Size} indicates the number of adjacent MR-context clips preserved.
}
\label{tab:sup_QCD_ratio_context}

\begin{adjustbox}{max width=\linewidth}
\begin{tabular}{c|c|ccc|cc}
\toprule
\textbf{Replace Ratio} & \textbf{Context Size} &
\multicolumn{3}{c|}{Moment Retrieval} & \multicolumn{2}{c}{Highlight Detection} \\ \cline{3-7}
 &  & R1@0.5 & R1@0.7 & mAP@0.5 & mAP & HIT@1 \\ 
\midrule

{}0.2 & {}0 & 64.12 & 50.54 & 64.88 & 40.78 & 67.40 \\ 

\rowcolor{gray!15}{}%
{}0.3 & {}0 & 64.45 & 50.92 & 65.37 & 41.38 & 68.02 \\ 

{}0.4 & {}0 & 64.04 & 51.23 & 64.58 & 41.21 & 67.76 \\ 
\midrule

\rowcolor{gray!20}{}%
{}0.3 & {}1 & \textbf{65.32} & \textbf{51.98} & \textbf{65.92} & \textbf{41.72} & \textbf{68.44} \\ 

{}0.3 & {}2 & 64.88 & 51.43 & 65.41 & 41.33 & 67.21 \\ 

{}0.3 & {}3 & 64.71 & 51.20 & 65.22 & 41.25 & 66.98 \\ 
\bottomrule
\end{tabular}
\end{adjustbox}
\end{table}

\subsection{Replacement Ratio and Context Preservation}

QCD additionally controls how much of the video is replaced and how much temporal context should be preserved.  
Table~\ref{tab:sup_QCD_ratio_context} shows that a \textbf{moderate replacement ratio} produces the strongest results:  
too small reduces augmentation diversity, while too large disrupts the video’s temporal structure.  
Similarly, preserving a \textbf{small boundary-adjacent context window} yields higher accuracy by maintaining essential temporal cues without overconstraining the augmentation process.

Together, these analyses indicate that QCD is effective when it operates \textbf{within an intermediate similarity band that excludes both trivial low-similarity backgrounds and high-similarity false negatives}, while using a moderate replacement ratio and a narrow boundary-adjacent context window. Under this configuration, QCD generates realistic yet alignment-consistent augmented samples, which in turn yields consistent gains in both moment retrieval and highlight detection performance.

\section{Additional Details about Context-invariant Boundary Discrimination (CBD)}

\subsection{Motivation}

Temporal boundaries constitute the most ambiguous and error-prone regions in moment retrieval. They occur at the interface between foreground and background, where semantic changes are abrupt and clip-level features exhibit significantly higher variance than within-moment interiors. Existing objectives—regression, bipartite matching, IoU-based losses, and rank-aware contrastive formulations—lack dedicated supervision for these transitional regions. This motivates a boundary-focused objective that (1) isolates boundary features, (2) enforces cross-view consistency under augmentation, and (3) strengthens discrimination against both temporally adjacent and semantically similar hard negatives.





\subsection{Boundary-focused Anchor Selection}

CBD is applied only at the start and end boundaries of each ground-truth moment. Since CBD relies on selecting a small set of temporal \textbf{anchor positions} at which contrastive consistency is enforced, we examine whether alternative anchor locations can serve the same role. To evaluate this, we compare three anchor-selection strategies: (1) \textit{All clips}, which treats every temporal index within GT moments as an anchor, (2) \textit{Center clips}, which uses only the central interior clips of the moment as anchors, and (3) our \textit{Boundary clips} formulation, which anchors exclusively at the start and end boundaries.

As shown in Table~\ref{tab:sup_boundary_all}, the \textbf{all-clip} variant yields the largest degradation across MR and HD metrics. When every index becomes an anchor, the contrastive objective over-constrains the feature space, suppressing natural temporal variation within the moment and interfering with regression losses that govern span prediction. 

The \textbf{center-clip} variant softens this effect, but center positions exhibit stable and less ambiguous semantics; consequently, they do not provide the hard contrastive signals needed to correct boundary-related localization errors.

In contrast, the \textbf{boundary-only} strategy consistently achieves the best performance. Boundary positions are precisely where temporal uncertainty is highest and where semantic transitions between foreground and background occur. Anchoring CBD at these positions introduces informative positives and challenging negatives, enabling the model to refine its boundary-sensitive representations. These results confirm that effective CBD requires \textbf{boundary-centric anchor selection}, and that applying contrastive supervision to interior or global positions is not beneficial.

\begin{table}[t]
\centering
\caption{Ablation on the scope of CBD anchor positions (QVHighlights val split).
\textbf{w/o CBD} denotes the model using QCD and CTE but without CBD.
}
\label{tab:sup_boundary_all}

\begin{adjustbox}{max width=\linewidth}
\begin{tabular}{c|ccc|cc}
\toprule
\textbf{Anchor choice} 
& R1@0.5 & R1@0.7 & mAP@0.5 & mAP & HIT@1 \\
\midrule

w/o CBD        
& 67.62 & 52.63 & 67.81 & 41.89 & 68.26 \\ \midrule

All clips        
& 65.22 & 48.12 & 63.56 & 39.78 & 66.77 \\

Center clips     
& 67.88 & 52.36 & 67.98 & 42.01 & 68.93 \\

\rowcolor{gray!15}
\textbf{Boundary clips (Ours)}   
& \textbf{69.61} & \textbf{54.84} 
& \textbf{67.97} & \textbf{43.47} & \textbf{70.40} \\

\bottomrule
\end{tabular}
\end{adjustbox}
\end{table}

\subsection{Boundary-IoU: A Boundary-centric Evaluation Metric}

Standard MR metrics evaluate overlap over the full moment span, which may remain high even when boundaries are misaligned. Since CBD explicitly targets boundary fidelity, we adopt \textbf{Boundary-IoU}, a metric designed to isolate boundary-localization quality.

Given a ground-truth moment \(M_{\text{GT}}=[s,e]\) and boundary width \(w\),
where \(s\) and \(e\) denote the start and end times (in seconds), and in 
QVHighlights these timestamps correspond to 2-second clip boundaries, we define:
\begin{align}
B_{\text{start}} &= [\, s,\; \min(s+w,\; e) \,], \\
B_{\text{end}} &= [\, \max(e-w,\; s),\; e \,].
\end{align}
Predicted boundary regions for $M_{\text{pred}}=[s',e']$ are defined analogously. Boundary-IoU is computed as:
\begin{equation}
\mathrm{Boundary\text{-}IoU} =
\frac{
\mathrm{IoU}(B_{\text{start}},B'_{\text{start}}) +
\mathrm{IoU}(B_{\text{end}},B'_{\text{end}})
}{2}.
\end{equation}
This metric focuses solely on the regions where boundary errors occur, providing a direct and sensitive measure of CBD's impact.
Table~\ref{tab:sup_cbd_boundary_iou} compares Boundary-IoU scores with and without CBD. CBD consistently improves both start- and end-boundary accuracy, confirming that it effectively models boundary-sensitive representations that are not captured by conventional IoU-based metrics.

\begin{table}[t]
\centering
\caption{\textbf{Boundary-IoU comparison before and after CBD.}
Evaluation conducted on QVHighlights val split with boundary width $w{=}2$.
Scores are computed for samples with whole-window IoU $\geq 0.7$.}
\label{tab:sup_cbd_boundary_iou}

\begin{adjustbox}{max width=\linewidth}
\begin{tabular}{l|ccc}
\toprule
Method & Start IoU & End IoU & Boundary IoU \\
\midrule
w/o CBD & 48.97 & 51.02 & 50.00 \\

\rowcolor{gray!10}
w/ CBD (Ours) & 
\textbf{52.54} {\scriptsize(\textcolor{red}{+7.29\% $\uparrow$})} &
\textbf{55.91} {\scriptsize(\textcolor{red}{+9.59\% $\uparrow$})} &
\textbf{54.26} {\scriptsize(\textcolor{red}{+8.52\% $\uparrow$})} \\
\bottomrule
\end{tabular}
\end{adjustbox}

\end{table}


\section{Additional Ablation Study of Context-enhanced Transformer Encoder (CTE)}

\textbf{The Context-enhanced Transformer Encoder (CTE)} is designed \textbf{to leverage the inherent continuity of video signals by aggregating information from neighboring clips}. Since adjacent video-clips often share motion cues and local semantics, incorporating multi-scale temporal receptive fields helps the model form more stable and context-aware representations. This is particularly beneficial for moment retrieval, where precise localization requires understanding both short-term transitions (e.g., motion boundaries) and long-range temporal context.

Table~\ref{tab:sup_CTE} presents an expanded ablation study exploring different combinations of receptive field sizes.
Each configuration such as \{5, 15, 75\} denotes the \textbf{temporal receptive field sizes} used at each block. 
 Smaller windows (e.g., 3 or 5) capture fine-grained motion patterns, whereas larger windows (e.g., 25 or 75) provide global temporal cues. Combining them yields consistent improvements over the baseline without CTE.
Especially, \textbf{CTE1 (\{5, 15, 75\})} demonstrates the best overall balance between retrieval accuracy and highlight detection performance. This configuration effectively integrates short-, mid-, and long-range temporal dependencies while maintaining minimal computational overhead. Consequently, we adopt \textbf{CTE1 as the default setting} in our model.

\begin{table}[t]
\centering
\caption{\textbf{Ablation on the Context-enhanced Transformer Encoder (CTE) in QVHighlights val split.}
Each CTE variant integrates different temporal receptive fields 
to capture multi-scale temporal context. 
CTE1 offers the best balance between accuracy and efficiency.}
\label{tab:sup_CTE}
\begin{adjustbox}{max width=\linewidth}
\begin{tabular}{l|ccc|cc}
\toprule
\multirow{2}{*}{\textbf{Variant}} & \multicolumn{3}{c|}{\textbf{Moment Retrieval}} & \multicolumn{2}{c}{\textbf{Highlight Detection}} \\ \cline{2-6}
& R1@0.5 & R1@0.7 & mAP@0.5 & mAP & HIT@1 \\
\midrule
\rowcolor{gray!15}
\textbf{Baseline (w/o CTE)} & 65.32 & 51.98 & 65.92 & 41.72 & 68.44 \\
\rowcolor{gray!10}
\textbf{CTE1 (Ours, \{5, 15, 75\})} & \underline{67.62} & \underline{52.63} & 67.81 & 41.89 & 68.26 \\
CTE2 (\{3, 5, 15\}) & 66.20 & 50.70 & 67.50 & 41.66 & 68.33 \\
CTE3 (\{3, 5, 25\}) & 66.53 & 51.34 & 67.75 & \underline{41.98} & \textbf{69.81} \\
CTE4 (\{5, 15, 25\}) & 66.84 & 51.82 & 67.71 & 41.81 & 68.59 \\
CTE5 (\{5, 25, 75\}) & 67.17 & 52.15 & 67.65 & 41.65 & 67.84 \\
CTE6 (\{3, 5, 15, 25\}) & 66.59 & 50.44 & \textbf{68.17} & \textbf{42.02} & \underline{68.91} \\
CTE7 (\{3, 5, 25, 75\}) & 66.53 & 50.89 & \underline{67.91} & 41.18 & 67.04 \\
CTE8 (\{3, 15, 25, 75\}) & \textbf{67.75} & 51.34 & 67.55 & 41.40 & 67.68 \\
CTE9 (\{5, 15, 25, 75\}) & 67.30 & 52.29 & 67.68 & 41.61 & 68.01 \\
CTE10 (\{3, 5, 15, 25, 75\}) & 67.56 & \textbf{52.82} & 67.46 & 41.40 & 66.20 \\
\bottomrule
\end{tabular}
\end{adjustbox}
\end{table}

\section{Additional Experimental Details}

\subsection{Implementation Details}
Following prior works \cite{lei2021Qvhighlights, radford2021learning, feichtenhofer2019slowfast}, 
we use pre-extracted multimodal features for all datasets. 
Video features are obtained from the pre-trained SlowFast network~\cite{feichtenhofer2019slowfast} 
and the CLIP vision encoder~\cite{radford2021learning}. 
Text queries are encoded using the corresponding CLIP text encoder. 
All features are provided at the clip level and kept frozen during training. 
We set the number of learnable queries in CTE to 100.
The hyperparameters $\alpha$ and $\beta$ for QCD are kept identical across all datasets.
All experiments are conducted on one NVIDIA A100 GPU (40GB memory, CUDA 11.8, Python 3.8).


\subsection{Evaluation Metrics}

We evaluate our model on three widely used benchmarks: QVHighlights, 
Charades-STA, and TACoS. Across these datasets, we follow standard 
protocols established in prior moment retrieval literature.

For \textbf{QVHighlights}, which includes both Moment Retrieval (MR) 
and Highlight Detection (HD) annotations, we report Recall@1 at IoU 
thresholds 0.5 and 0.7, mAP@0.5, and the average mAP computed over 
IoU thresholds from 0.5 to 0.95 with a step size of 0.05. For HD, 
we additionally report the HIT@1 metric, which measures whether the 
highest-scoring clip corresponds to a ground-truth highlight. 
This combination of metrics captures retrieval accuracy, temporal 
localization precision, and highlight scoring quality.

For \textbf{Charades-STA} and \textbf{TACoS}, we follow prior work and 
evaluate performance using Recall@1 at IoU thresholds 0.5 and 0.7. 
These datasets focus purely on moment retrieval without highlight labels, 
making R1-based localization accuracy the standard evaluation measure.
This consistent metric set provides a comprehensive view of retrieval 
correctness, boundary alignment quality, and highlight detection 
performance across the different datasets.

\section{Robustness to Spurious Correlations}

A core claim of our framework is that CVA learns context-invariant representations rather than relying on spurious correlations between queries and static backgrounds. To directly validate this, we adopt the \textit{target-masked} diagnostic protocol from TD-DETR~\cite{zhou2025devil}:
(i) \textbf{Random masking} replaces GT-moment clips with noise matching the original feature statistics, and
(ii) \textbf{Zero masking} removes GT content entirely by setting features to zero.
If a model relies on background context rather than the actual target moment, it will still produce high retrieval scores under these masking conditions.

As shown in Table~\ref{tab:sup_spurious}, our model consistently achieves lower spurious scores than TD-DETR under both protocols.
Notably, under \textbf{Zero masking}, CVA reduces spurious R1@0.7 from 21.23 to \textbf{7.16} (a \textbf{66\%} reduction), demonstrating that our model genuinely relies on the target moment content rather than contextual bias. This confirms that the combination of QCD augmentation and CBD loss effectively enforces context-invariant learning.

\begin{table}[t]
\centering
\caption{Spurious correlation diagnostic on QVHighlights \textit{val} split. \textbf{Random} replaces GT clips with noise matching original feature statistics; \textbf{Zero} sets GT features to zero. Lower values indicate less reliance on background context.}
\label{tab:sup_spurious}
\begin{adjustbox}{max width=\linewidth}
\begin{tabular}{c|c|cc|cc}
\toprule
\textbf{Mask mode} & \textbf{Method}
& \multicolumn{2}{c|}{\textbf{Spurious R1} $\downarrow$}
& \multicolumn{2}{c}{\textbf{Spurious mAP} $\downarrow$} \\
\cline{3-6}
& & R1@0.7 & R1@0.9 & @0.75 & Avg. \\
\midrule
\multirow{2}{*}{Random}
& TD-DETR & 2.45 & 1.03 & 3.18 & 3.82 \\
& \cellcolor{gray!15}\textbf{Ours} & \cellcolor{gray!15}\textbf{2.39} & \cellcolor{gray!15}\textbf{0.84} & \cellcolor{gray!15}\textbf{2.70} & \cellcolor{gray!15}\textbf{3.17} \\
\midrule
\multirow{2}{*}{Zero}
& TD-DETR & 21.23 & 14.00 & 21.35 & 20.93 \\
& \cellcolor{gray!15}\textbf{Ours} & \cellcolor{gray!15}\textbf{7.16} & \cellcolor{gray!15}\textbf{5.16} & \cellcolor{gray!15}\textbf{7.53} & \cellcolor{gray!15}\textbf{7.48} \\
\bottomrule
\end{tabular}
\end{adjustbox}
\end{table}

\section{Qualitative Results and Analysis}

To further compare temporal grounding behavior across models, we present qualitative examples in Fig.~\ref{fig:negative_false}. The first example represents a challenging scenario in which the camera frequently focuses on food rather than the person cooking. Because the visual evidence for the target action appears only intermittently across clips, accurate localization requires integrating both \textbf{short-range temporal transitions} (e.g., brief motion onsets or local dynamics) and \textbf{long-range temporal structure} (e.g., scene progression and repeated contextual cues). CG-DETR fails to identify the target moment, and TD-DETR captures only a marginal portion with limited alignment. In contrast, our model closely matches the ground-truth interval. This robustness arises from the combined contributions of \textbf{QCD} (which prevents semantic contamination during augmentation), \textbf{CTE} (which enhances multi-scale temporal reasoning across clips), and  
\textbf{CBD} (which sharpens boundary discrimination). These components collectively enable accurate grounding even when \textbf{clip-level appearance cues are weak, unreliable, or partially missing}.

In the second example, both CG-DETR and TD-DETR activate a number of \textbf{false-positive} segments that are not semantically related to the query. Our method suppresses these spurious responses and localizes the intended region more precisely, demonstrating stronger discriminative ability under complex and visually distracting background conditions.

The third example consists of a sequence of short, rapidly transitioning actions. Our model accurately resolves these \textbf{fine-grained temporal boundaries}, whereas CG-DETR merges them into a single coarse segment and TD-DETR fails to capture the initial action entirely. This highlights the effectiveness of our boundary-sensitive design in handling 
dense and fast-changing temporal structures.

Finally, as shown in Fig.~\ref{fig:ours}, the predicted saliency distribution is concentrated sharply within the ground-truth interval. This provides an interpretable visualization of how our model identifies relevant temporal cues while suppressing irrelevant clips.

\begin{figure*}[!t]
\centering
\includegraphics[width=0.95\textwidth]{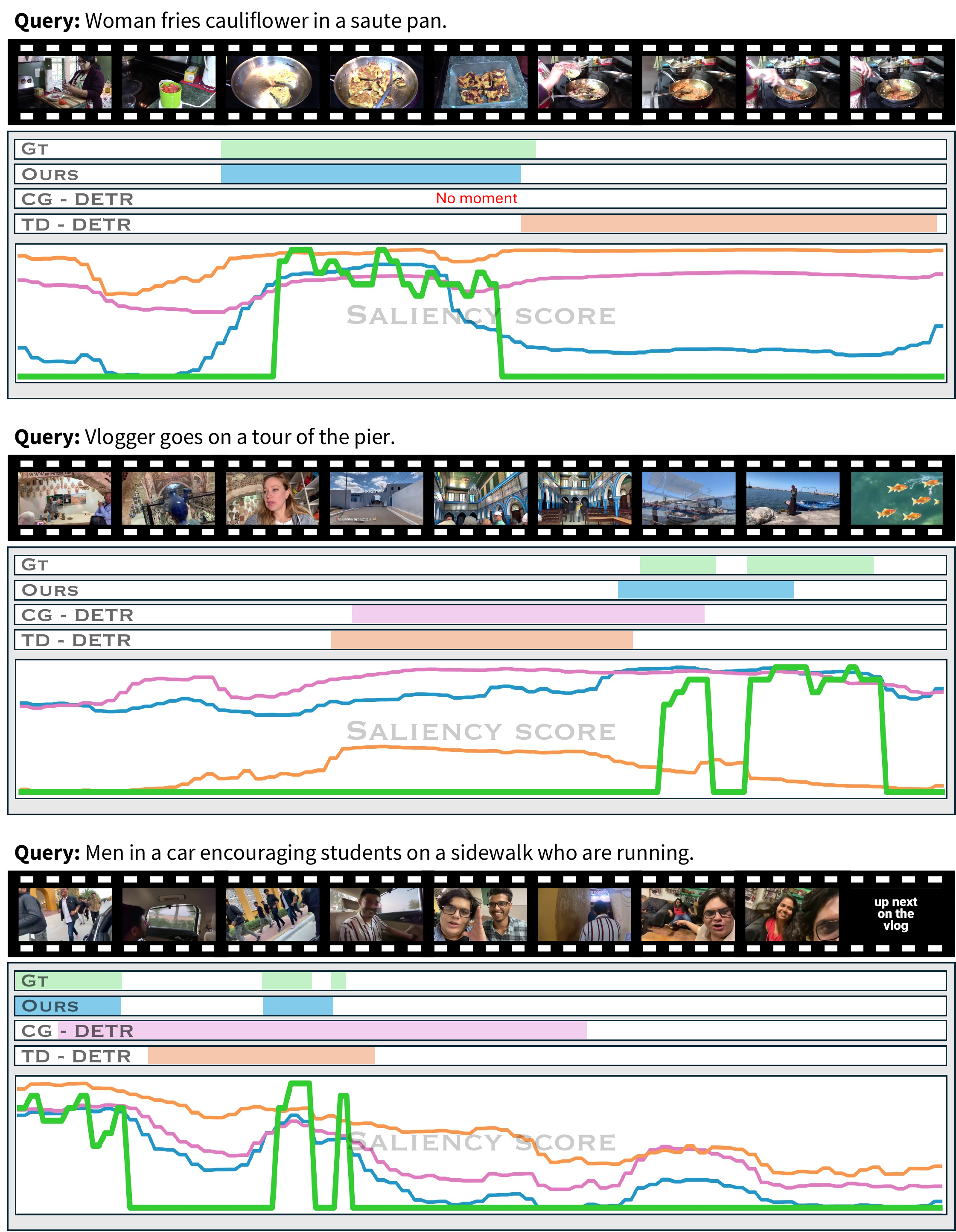}
\caption{\textbf{Qualitative comparison with CG-DETR and TD-DETR on QVHighlights.}
Our model aligns more accurately with ground-truth moments, 
reduces false positives, and resolves fine-grained temporal transitions more effectively. 
Saliency responses are also better concentrated within ground-truth intervals,
reflecting improved moment discrimination.}
\label{fig:negative_false}
\end{figure*}

\begin{figure*}[!t]
\centering
\includegraphics[width=0.95\textwidth]{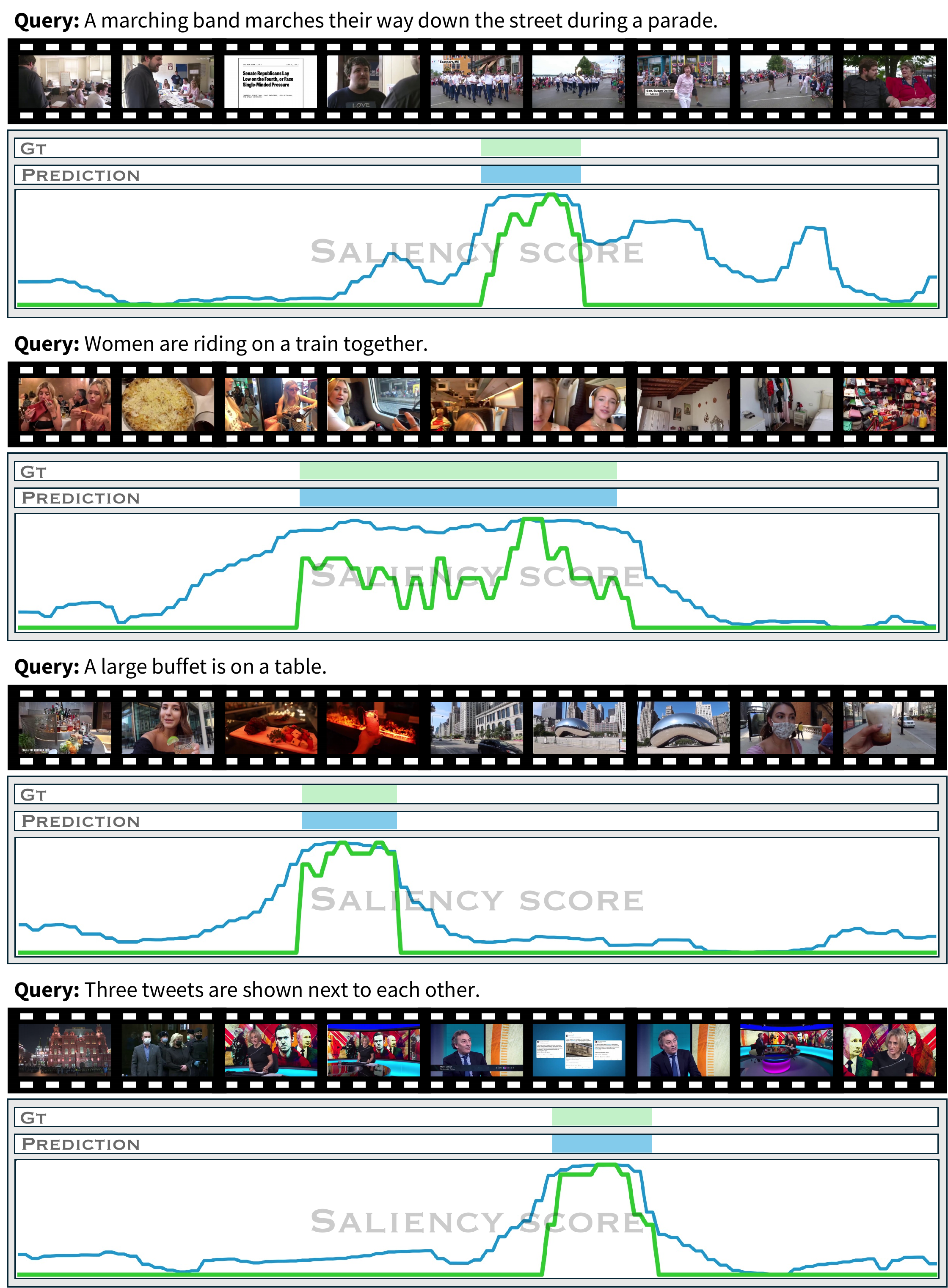}
\caption{Qualitative results of our method.}
\label{fig:ours}
\end{figure*}


\end{document}